\documentclass[letterpaper]{article}
\usepackage{aaai}
\usepackage{times}
\usepackage{helvet}
\usepackage{courier}
\usepackage{graphicx}

\begin{document}
%
\title{SkyNet: A Champion Model for DAC-SDC on Low Power Object Detection}
\author{Xiaofan Zhang$^{1,3}$, Cong Hao$^1$, Haoming Lu$^1$, Jiachen Li$^1$, Yuhong Li$^1$, Yuchen Fan$^1$, Kyle Rupnow$^3$, \vspace{+6pt}\\  \Large \textbf{Jinjun Xiong$^{2,1}$, Thomas Huang$^1$, Honghui Shi$^{2,1}$, Wen-mei Hwu$^1$, Deming Chen$^{1,3}$} \vspace{+6pt}\\
$^1$IBM-Illinois Center for Cognitive Computing Systems Research (C$^3$SR), Univ. of Illinois at Urbana-Champaign\vspace{+4pt}\\
$^2$IBM T. J. Watson Research Center\vspace{+4pt}, 
$^3$Inspirit IoT, Inc\vspace{+2pt}\\ 
\textit{\{xiaofan3, congh, hl36, jiachenl, leeyh, yuchenf4, t-huang1, w-hwu, dchen\}@illinois.edu,} \vspace{+4pt}\\\textit{jinjun@us.ibm.com,  honghui.shi@ibm.com, kyle.rupnow@inspirit-iot.com}
}
\vspace{+4pt}
\nocopyright
\maketitle
\vspace{+14pt}
\begin{abstract}
\begin{quote}
Developing artificial intelligence (AI) at the edge is always challenging, since edge devices have limited computation capability and memory resources but need to meet demanding requirements, such as real-time processing, high throughput performance, and high inference accuracy. To overcome these challenges, we propose SkyNet, an extremely lightweight DNN with 12 convolutional (Conv) layers and only 1.82 megabyte (MB) of parameters following a bottom-up DNN design approach. SkyNet is demonstrated in the 56th IEEE/ACM Design Automation Conference System Design Contest (DAC-SDC), a low power object detection challenge in images captured by unmanned aerial vehicles (UAVs). SkyNet won the first place award for both the GPU and FPGA tracks of the contest: we deliver 0.731 Intersection over Union (IoU) and 67.33 frames per second (FPS) on a TX2 GPU and deliver 0.716 IoU and 25.05 FPS on an Ultra96 FPGA.

\end{quote}
\end{abstract}

\section{Introduction}
Edge AI applications not only require high inference accuracy of deep neural networks (DNNs), but also require more aggressive inference speed (e.g, low latency for real-time response and high throughput for supporting streaming inputs) and efficiency (e.g, low power and energy consumption for less heat and longer battery life). These applications are in urgent need of hardware acceleration using energy efficient edge devices, such as embedded GPUs \cite{franklin2017nvidia} and FPGAs \cite{zhang2018dnnbuilder}. 

As both inference accuracy and efficiency are the key factors to distinguish good solutions,  software algorithms (DNNs) need to be compact and hardware friendly; otherwise, it is impossible to overcome the hardware limitations under various edge-computing scenarios. 
To demonstrate our DNN design, 
we participated in the DAC-SDC, which features a low power object detection challenge and asks for novel object detection solutions on the resource-constrained embedded hardware platforms \cite{xu2018dac}. This challenge aims at a single object detection task from real-life UAV applications and requires comprehensive evaluations among detection accuracy, throughput, and energy consumption on two targeted embedded platforms (Nvidia TX2 GPU and Xilinx Ultra96 FPGA). 

\begin{figure}[t]
\centering
\includegraphics[width=0.48\textwidth]{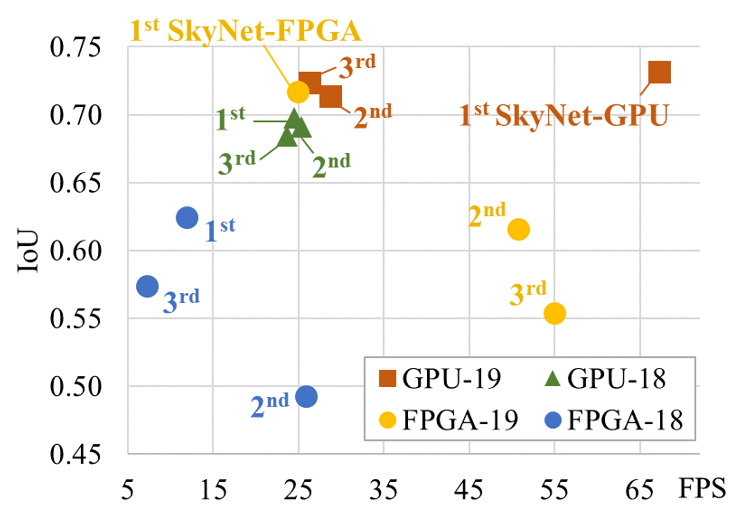}
\vspace{-10pt}
\caption{DAC-SDC Top-3 design comparison of inference throughput (FPS) and inference accuracy (IoU) on edge devices for last two years. The GPU track of DAC-SDC'19 and '18 both target TX2 GPU with the same object detection task. For the FPGA track, the designs from this year (FPGA-19) use Ultra96 FPGA while the designs from last year (FPGA-18) use Pynq-Z1 FPGA. Our designs (SkyNet-GPU in GPU track and SkyNet-FPGA in FPGA track) used the proposed SkyNet and demonstrated large advantages in the respective tracks.}
\vspace{-4pt}
\label{fig:rst_comp}
\end{figure}

\begin{figure*}[t]
\centering
\includegraphics[width=0.90\textwidth]{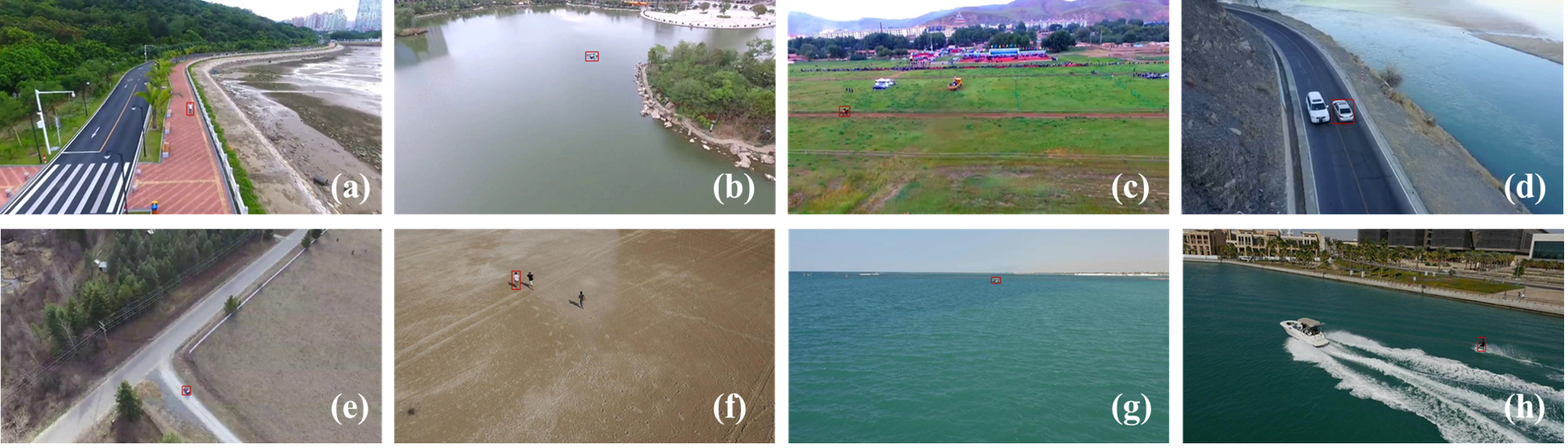}
\vspace{-8pt}
\caption{Examples from the training dataset with red color bounding boxes in main categories as rider (a), drone (b), horse rider (c), car (d, e), person (f), boat (g), and wakeboard (h). }
\vspace{-4pt}
\label{fig:dataset}
\end{figure*}

\begin{table*}[t]
\vspace{-0pt}
\caption{2018 DAC-SDC Top-3 entries from both GPU and FPGA tracks. These designs follow a top-down DNN design approach starting from pre-defined models and then apply network compression techniques including: \textcircled{\tiny{1}} input resizing, \textcircled{\tiny{2}} network pruning, \textcircled{\tiny{3}} low-precision data representation, and \textcircled{\tiny{4}} TensorRT.}
\label{tab:dacsdc2018}
\begin{center}
\begin{small}
\newcommand{\tabincell}[2]{\begin{tabular}{@{}#1@{}}#2\end{tabular}}
\begin{tabular}{|c|c|c|c|c|c|}
\hline
\textbf{  Rank  } & \textbf{GPU-Track} & \textbf{Predefined DNN} & \textbf{    IoU    } & \textbf{FPS} & \textbf{  Compression strategies  }\\
\hline
1st & \tabincell{c}{ICT-CAS \cite{ICT-CAS}} & Tiny YOLO & 0.698 & 24.55 & \textcircled{\tiny{1}} \textcircled{\tiny{2}} \textcircled{\tiny{3}} \textcircled{\tiny{4}} \\ \hline
2nd & \tabincell{c}{DeepZ \cite{DeepZ}}& Tiny YOLO & 0.691 & 25.30 & Not clear \\ \hline
3rd & \tabincell{c}{SDU-Legend \cite{SDU-Legend}}& YOLOv2 & 0.685 & 23.64 & \textcircled{\tiny{1}} \textcircled{\tiny{2}} \textcircled{\tiny{3}} \\ 
\hline
\textbf{Rank} & \textbf{FPGA-Track} & \textbf{Predefined DNN} & \textbf{IoU} & \textbf{FPS} & \textbf{Compression strategies} \\
\hline
1st & \tabincell{c}{TGIIF \cite{TGIIF}} & SSD & 0.624 & 11.96 & \textcircled{\tiny{1}} \textcircled{\tiny{2}} \textcircled{\tiny{3}} \\ \hline
2nd & \tabincell{c}{systemsETHZ \cite{systemsETHZ}} & SqueezeNet & 0.492 & 25.97 & \textcircled{\tiny{1}} \textcircled{\tiny{2}} \textcircled{\tiny{3}} \\ \hline
3rd & \tabincell{c}{iSmart2 \cite{iSmart2}} & MobileNet & 0.573 & 7.35 & \textcircled{\tiny{1}} \textcircled{\tiny{2}} \textcircled{\tiny{3}} \\ 
\hline
\end{tabular}
\end{small}
\end{center}
\end{table*}

We address the task in DAC-SDC by proposing SkyNet --- a lightweight object detection DNN developed by a novel bottom-up design approach.
The main contributions of this paper are summarized as follows:

\begin{itemize}
    \item We summarize the designs proposed by the top-3 winners of DAC-SDC 2018 (both GPU and FPGA tracks) to locate the potential obstacles for achieving better detection accuracy and hardware efficiency.
    \item We develop SkyNet using a bottom-up design approach with comprehensive awareness of the hardware limitations. In addition, three more features are added including: feature map bypassing, reordering and ReLU6 (instead of ReLU).
    \item We deploy the proposed SkyNet on both TX2 GPU and Ultra96 FPGA and achieve the highest IoU and total score in DAC-SDC 2019, winning the first place award for both GPU and FPGA tracks (Figure \ref{fig:rst_comp}). 
\end{itemize}

\section{DAC-SDC}
This year's DAC-SDC launches an object detection challenge for images taken from drones with comprehensive evaluation system considering design accuracy, throughput, and energy consumption. The goal of this competition is to provide unified edge-platforms to develop and compare state-of-the-art object detection system design.

\subsection{Targeted UAV applications}
DAC-SDC targets the single object detection, which is one of the most important tasks in real-life UAV applications. It considers the most appropriate needs of UAV applications, such as capability of real-time processing, energy efficiency, and detection accuracy.
To better reflect real-life challenges, images of the dataset are captured by UAVs and they are provided by the drone manufacturer DJI. The whole dataset is divided by two parts: the training dataset (100,000 images with objects of interest across 12 main categories and 95 sub-categories) and the hidden test set for official evaluation (50,000 images that only the contest organizers could access) \cite{DAC-SDC-dataset}. Examples of the training dataset are shown in Figure \ref{fig:dataset} and most of these objects are very small and challenging to detect.

\subsection{Previous winning design summary}
\begin{figure}[t]
\centering
\includegraphics[width=0.49\textwidth]{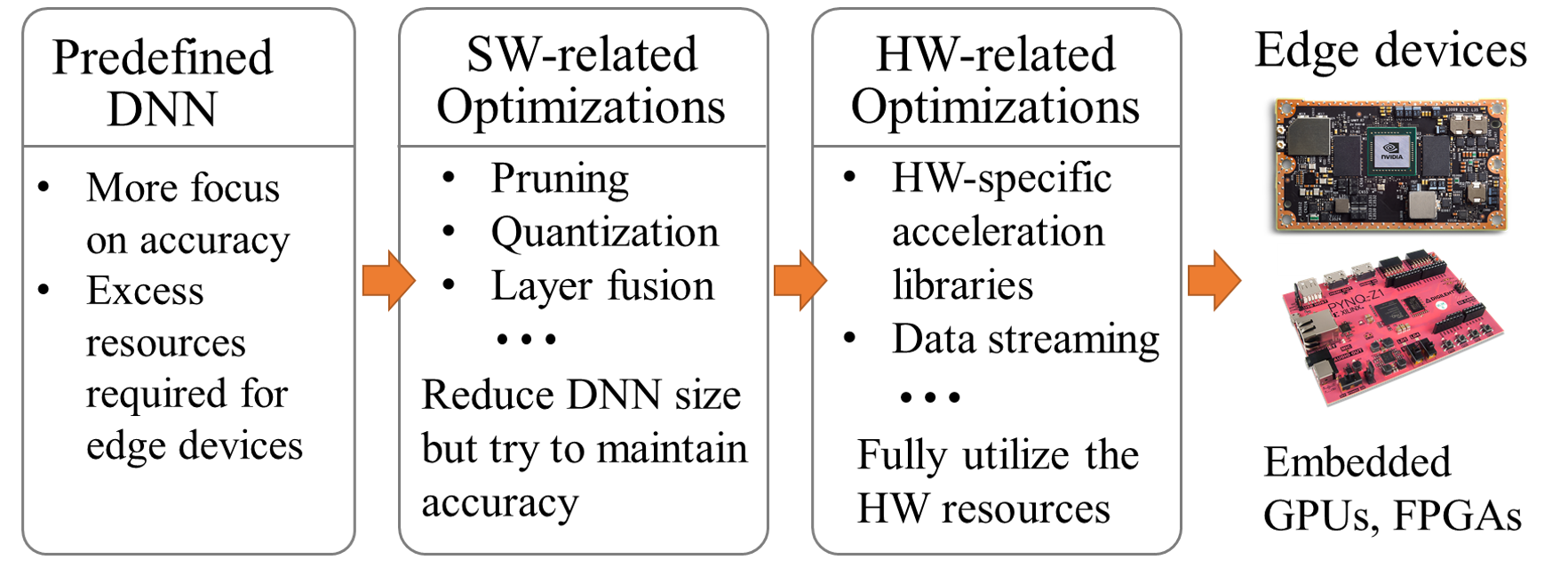}
\vspace{-8pt}
\caption{A widely-used top-down DNN design approach for DNN development and deployment on resource-constrained hardware platforms on the edge.}
\vspace{-4pt}
\label{fig:top-down}
\end{figure}

By examining the top-3 entries from both GPU and FPGA tracks, we notice that all of these designs share a similar top-down DNN design approach as shown in Figure \ref{fig:top-down}. They first adopt well-known DNNs which show outstanding accuracy in similar tasks (such as YOLO \cite{redmon2016you} and SSD \cite{liu2016ssd} for object detection) and then apply optimization techniques on the software and hardware sides, trying to compress the network size to fit into edge devices. Since these well-behaved DNNs are originally accuracy-orientated on general computing platforms such as desktop and server GPUs, they may not perform well on resource limited edge devices, which easily introduce quality degradation to final results.

We summarize the top-3 designs of DAC-SDC 2018 in Table \ref{tab:dacsdc2018} with their predefined DNNs as well as the compression strategies they used. All of the GPU teams start from YOLO, and result in IoU lower than 0.7 and throughput around 25 FPS. To compress the original DNNs, participants employ input resizing to lower the computational complexity and network pruning to reduce the unnecessary network connections. They also use half-precision floating precision (16-bit) instead of 32-bit to improve throughput. In the FPGA track, participants are required to conduct more aggressive network compression because of the tighter hardware resource budget on the targeted FPGA. In these designs, DNN parameters are greatly reduced by shrinking the DNN depth (the number of DNN layers) and width (the number of channels for each layer) of the predefined DNNs. Meanwhile, the bit-width of the DNN parameters are quantized to 8 bits or even 1 $\sim$ 2 bits. 

\begin{figure*}[t]
\centering
\includegraphics[width=0.80\textwidth]{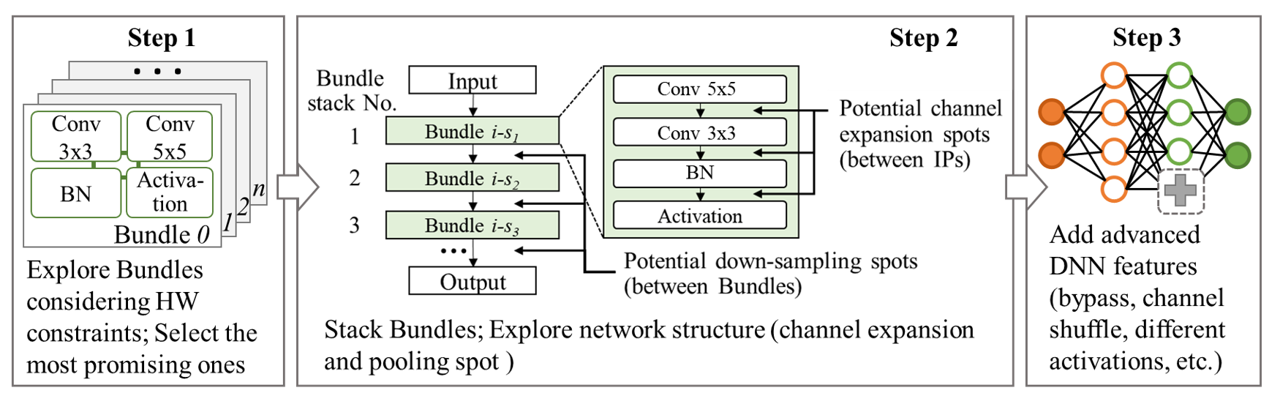}
\vspace{-8pt}
\caption{The bottom-up DNN design flow we adopt from \cite{hao2019fpga} for SkyNet design. Without relying on certain predefined DNNs, we start building our own network from scratch following three steps: 1) Bundle selection, 2) network search, and 3) feature addition. }
\vspace{-12pt}
\label{fig:bottom-up-flow}
\end{figure*}

\section{Motivations}
Since all top-3 GPU teams in 2018 have fully investigated the potential of YOLO, we start our preliminary investigations on SSD, another popular DNN candidate for object detection. By following the top-down DNN design approach (Figure \ref{fig:top-down}) as all teams have done, we pick two predefined backbone networks (VGG16 \cite{simonyan2014very} and MobileNet \cite{howard2017mobilenets}) for feature extraction with input size 3$\times$360$\times$640. We use the DAC-SDC training dataset (containing 100K images) for DNN training and the officially provided 1000 images for validation. After sufficient training, the accuracy results are 0.70 and 0.66 IoU using VGG16 and MobileNet backbone, respectively. Without network compression, these two SSD models can only run 15 FPS (with VGG16) and 24 FPS (with MobileNet) on a desktop GPU (Nvidia 1080Ti). Still, great efforts need to be made for adapting them to the targeted embedded platforms. 

Following the top-down DNN design flow, we face two essential challenges preventing us for reaching better solutions with both higher inference accuracy and faster inference speed:

\begin{itemize}
    \item[] 1) Similar inference speedup (with similar DNN compression ratio) but vastly different accuracy.
    \item[] 2) Uncertain inference accuracy variation for a given task.
\end{itemize}

For the first challenge, the underlying factor is the different sensitivities of DNN configurations regarding inference accuracy and hardware performance. It is hard to make a perfect balance since negligible changes in DNN model may cause huge differences during its hardware deployment and vice versa, resulting in difficult trade-off between inference accuracy and hardware performance.

An example regarding the compression ratio of AlexNet \cite{krizhevsky2012imagenet} and its inference accuracy on ImageNet dataset \cite{deng2009imagenet} is given in \cite{zhang2019bi}. By using data quantization, the memory footprints of parameter and feature map start shrinking to enable better inference throughput. However, the accuracy trends of the designs with parameter and feature map quantization vary significantly; obviously, the precision of feature map contributes more to the inference accuracy than the parameters. To overcome the first challenge, we need to study the accuracy and speed sensitivity of each component of the DNN before network compression.

For the second challenge, the accuracy upper bound on a given task is very difficult to determine. An experiment to evaluate the accuracy variation on DAC-SDC dataset is presented in \cite{zhang2019bi}. With the fixed back-end bounding box regression part, the well-known DNNs (including VGG16 and ResNet \cite{he2016deep}) are respectively adopted as the backbone to get accuracy results in targeted dataset. However, there are no clear clues regarding the network size and inference accuracy even for the same architecture as ResNet-18, -32, and -50. It is not easy to select a promising predefined model for a given task following the top-down design approach. 

\section{New Approach: Bottom-up DNN Design}
Due to the two challenges mentioned above, we try a different design direction --- a bottom-up DNN design strategy, and start building DNN from scratch.
We adopt the DNN design method from \cite{hao2019fpga} and propose a hardware-oriented DNN model with adequate understanding of hardware constraints, hoping that our design can balance well between inference accuracy and performance on the edge devices. The overall bottom-up design flow is shown in Figure \ref{fig:bottom-up-flow}, 

\subsection{Step 1: Bundle construction}
The proposed design flow starts with constructing the hardware-aware basic building blocks called Bundles. In our definition, a Bundle is a set of sequential DNN layers, which can be repeatedly stacked and eventually construct DNNs.
To guarantee that Bundles have full understandings of hardware constraints, we use analytical models for performance and resource estimation of Bundles, so that we can select Bundles according to their hardware performance.

In the first step, we enumerate the DNN components from different essential DNN layer types (such as Conv, Pooling, activation layer, etc.) and assemble them into Bundle $0 \sim n$.  Since our design needs to target both GPU and FPGA tracks, we use the resource constraints from FPGA side (more restrictive compared to GPU) to evaluate the hardware performance (e.g., inference latency) for each Bundle. During implementation, all DNN components inside a Bundle are instantiated on the targeted hardware, which means the larger Bundle (with more DNN components) results in higher resource overhead and longer latency, and it is less likely to be selected.

In order to get each Bundle's potential accuracy contribution, we build a DNN sketch with fixed front-end and back-end structures, and insert one Bundle (with replications) in the middle each time. In our case, the front-end and back-end are the input resizing and bounding box regression, respectively. Each DNN is quickly trained for 20 epochs to get its accuracy. Since each Bundle has its own characteristics regarding latency, accuracy and resource overhead, the most promising Bundles with satisfied speed and best relative accuracy are selected for the next step.

\subsection{Step 2: Network search}
In step 2, we perform DNN structure search.
The required inputs include the initial DNNs (built by stacking selected Bundles), the latency target $Lat_{targ}$, the acceptable latency tolerance $\epsilon$, and the resource constraint $Res_{max}$, while the outputs are the generated DNNs. 
We use stochastic coordinate descent (SCD) to update three variables related to the DNN structure: the number of Bundle replications; down-sampling configuration between Bundles; and channel expansion configuration. Assuming the achieved latency and resource overhead of the generated DNN are $Lat$ and $Res$, the objective of using SCD is  $|Lat_{targ} - Lat| < \epsilon$ and $Res < Res_{max}$. These three variables contribute three coordinates and the SCD algorithm picks one coordinate randomly every iteration and updates DNN structure along that direction \cite{hao2019fpga}. As we specify the latency target at the very beginning, the generated DNNs after network search favor lightweight structures for the targeted hardware platforms. 


In addition, we have made two contributions to boost the DNN search efficiency. First, during step 1, we already select the promising Bundles regarding latency, accuracy, and resource overhead. Since such exploration has been done in step 1, the search process in step 2 is not overwhelming. Second, we limit the DNN design space for faster search, to only explore the number of layers, channel expansion insertion locations, and the pooling opportunities. As a result, the generated DNNs are more structured with traditional stacked network architecture. Such regular network structures help boost the hardware efficiency while deploying on the edge devices.

\subsection{Step 3: Feature addition}
In step 3, we add more advanced DNN features when hardware resources are allowed to further tailor the generated network for targeted tasks. For DAC-SDC, since most of the objects to be detected are small, we add a bypass with feature map reordering in SkyNet to strengthen the capability of small object detection. To enhance the hardware efficiency, we replace the ReLU with ReLU6. More discussions are provided in the next section. 

\begin{figure*}[t!]
\centering
\vspace{-8pt}
\includegraphics[width=0.80\textwidth]{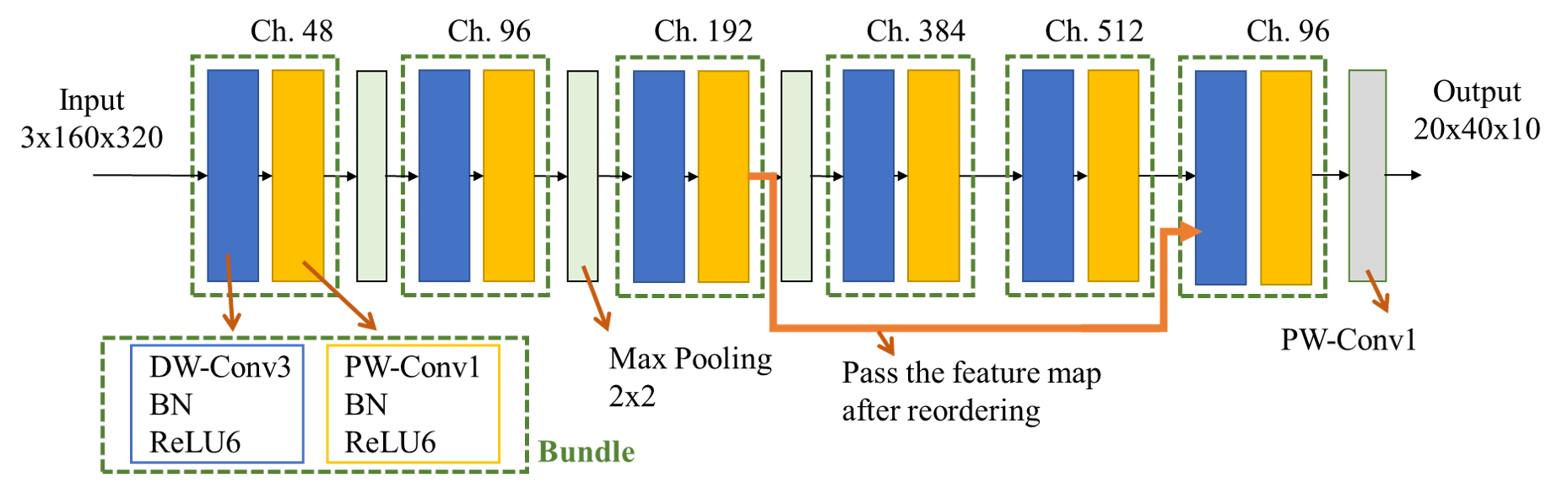}
\vspace{-8pt}
\caption{SkyNet architecture (version C in Table \ref{tab:propose-dnn}) generated by stacking six of the selected Bundle (circled by green dash line) with DNN components as: DW-Conv3, PW-Conv1, BN, and ReLU6. The number of output channels is listed on top of each Bundle denoted as Ch. Three 2$\times$2 pooling layers are inserted. The bypass is highlighted in orange, which passes feature maps generated by the third Bundle directly to the last Bundle. The feature map reordering is also performed along with the bypass.}
\vspace{-12pt}
\label{fig:SkyNet}
\end{figure*}

\section{SkyNet}
The main idea of building SkyNet is to explore DNN design following the bottom-up approach to deliver AI capabilities on resource-constrained edge devices.

\subsection{SkyNet architecture}
The bundle we selected is a combination of 3$\times$3 depth-wise convolutional layer (DW-Conv3), 1$\times$1 point-wise convolutional layer (PW-Conv1), batch normalization layer (BN), and rectified linear unit 6 (ReLU6). By repeatedly stacking this Bundle, we generate the backbone network architecture A in Table \ref{tab:propose-dnn}. Advanced features are added afterwards shown in architecture B and C (Table \ref{tab:propose-dnn}). For the back-end of SkyNet, we adapt the YOLO back-end by removing the classification output and use two anchors for bounding box regression.

\begin{table}[t]
\vspace{-8pt}
\caption{The detailed SkyNet structure. The number of channels of each layer is shown in the bracket. Each convolutional layer except the last one is followed by a batch normalization and a ReLU6 (omitted for conciseness). 
}
\footnotesize   
\label{tab:propose-dnn}
\begin{center}
\begin{tabular}{|c|c|c|}
\hline
\multicolumn{3}{|c|}{Configurations of SkyNet} \\ \hline
A   & B   & C  \\ \hline
\multicolumn{3}{|c|}{input (3$\times$160$\times$360 color image)} \\ \hline
\multicolumn{3}{|c|}{\begin{tabular}[c]{@{}c@{}}DW-Conv3 (3) \\ PW-Conv1 (48)\end{tabular}} \\ \hline
\multicolumn{3}{|c|}{2$\times$2 max-pooling} \\ \hline
\multicolumn{3}{|c|}{\begin{tabular}[c]{@{}c@{}}DW-Conv3 (48) \\ PW-Conv1 (96)\end{tabular}} \\ \hline
\multicolumn{3}{|c|}{2$\times$2 max-pooling} \\ \hline
\multicolumn{3}{|c|}{\begin{tabular}[c]{@{}c@{}}DW-Conv3 (96) \\ PW-Conv1 (192) \\ \textit{\textbf{$[$Bypass Start$]$ FM Reordering}} (768) \end{tabular}} \\ \hline
\multicolumn{3}{|c|}{2$\times$2 max-pooling} \\ \hline
\multicolumn{3}{|c|}{\begin{tabular}[c]{@{}c@{}}DW-Conv3 (192) \\ PW-Conv1 (384) \\ \end{tabular}} \\ \hline
\multicolumn{3}{|c|}{\begin{tabular}[c]{@{}c@{}}DW-Conv3 (384) \\ PW-Conv1 (512) \\ \end{tabular}} \\ \hline

\begin{tabular}[c]{@{}c@{}}
PW-Conv1 (10) \\ \end{tabular} & \begin{tabular}[c]{@{}c@{}} \textit{\textbf{$[$Bypass End$]$}} \\ \textit{\textbf{FM Concatenated}} \\ DW-Conv3 (512+768) \\ PW-Conv1 (48) \\ PW-Conv1 (10) \\ \end{tabular} & \begin{tabular}[c]{@{}c@{}} \textit{\textbf{$[$Bypass End$]$}} \\ \textit{\textbf{FM Concatenated}}\\ DW-Conv3 (512+768) \\ PW-Conv1 (96)\\PW-Conv1 (10) \\
\end{tabular} \\ \hline
\multicolumn{3}{|c|}{Back-end for bounding box regression}  \\ \hline
\end{tabular}
\end{center}
\vspace{-14pt}
\end{table}

\begin{figure}[t]
\centering
\includegraphics[width=0.35\textwidth]{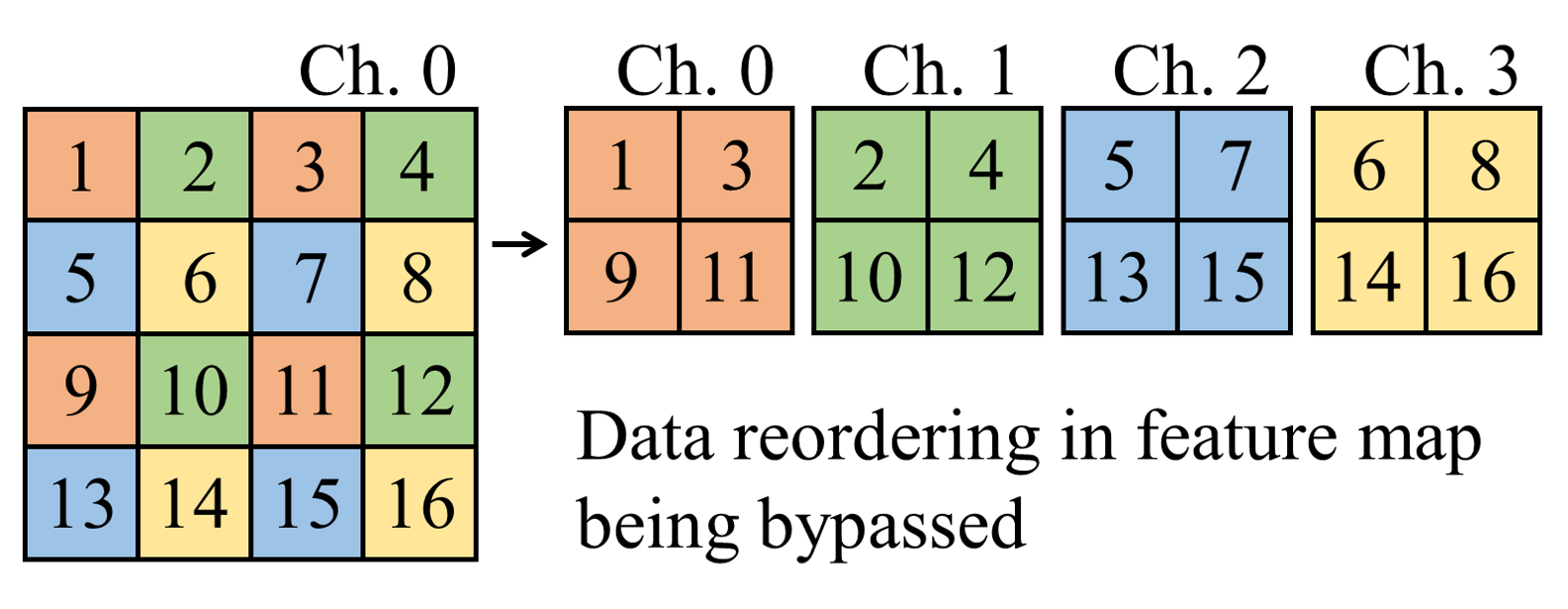}
\vspace{-4pt}
\caption{Feature map reordering from $1\times4\times4$ to $4\times2\times2$ with shrunken width and height but expanded number of channels. There is no information loss compared to traditional pooling. In addition, this reorder pattern also ensures larger receptive field.}
\vspace{-4pt}
\label{fig:reorder}
\end{figure}

\subsection{Feature map bypassing and reordering}

\begin{figure}[t]
\centering
\includegraphics[width=0.5\textwidth]{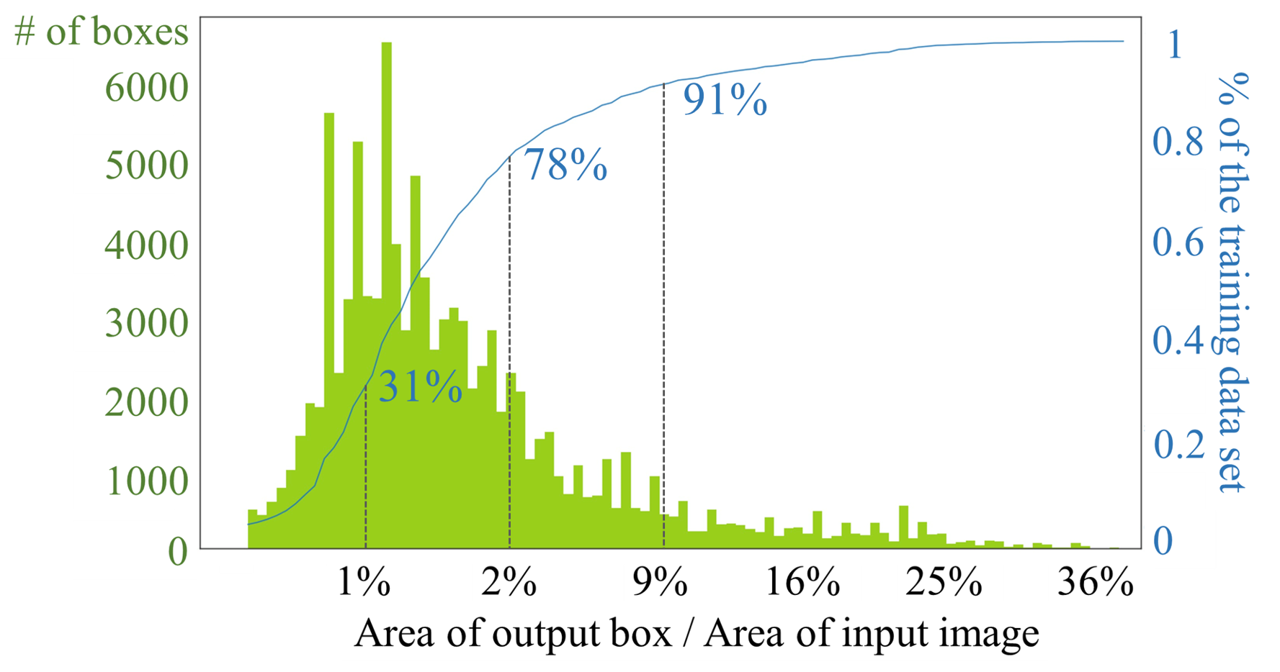}
\vspace{-22pt}
\caption{The distribution of bounding box relative size in DAC-SDC training dataset. We capture the bounding box relative size by computing the ratio of output bounding box size divided by the input image size. The green bars show the ratio distribution, and the blue curve shows the corresponding cumulative distribution.}
\vspace{-16pt}
\label{fig:data_analysis}
\end{figure}

By examining the competition training data, we keep a record of the size ratio between the output bounding box and the input image and present a distribution diagram in Figure \ref{fig:data_analysis}. It clearly shows that 91\% of the objects to be detected in DAC-SDC dataset are less than 9\% of the original input image size and 31\% of them are even smaller than 1\% of the input image size. It means the majority of objects in this dataset can be considered as small objects and we need to propose a DNN accordingly.

We add feature map bypassing and reordering to enhance the ability of small object detection. The bypass helps to keep small object features until the deeper part of the DNN without being diluted by the pooling layers. Also, it is beneficial to have multiple feature maps (from different layers) before generating the bounding boxes. Since the bypass crosses a pooling layer (highlighted in Figure \ref{fig:SkyNet}), we use reordering (shown in Figure \ref{fig:reorder}) to align the size of original feature map (generated by the fifth Bundle) and the bypassed one without losing valuable features. 

\subsection{ReLU6}
The other feature we use to improve hardware efficiency is ReLU6, which is a activation function with compressed output range $[0,6]$. Since ReLU6 asks for much smaller data range compared to the original ReLU ($[0,+ \infty)$), less bits are required to represent activations, such as using lower-precision floating point in embedded GPUs, or fixed-point data type in embedded FPGAs.

\subsection{Training}

We train the SkyNet in an end-to-end fashion using multi-scale training with the learning rate starting from 1e-4 to 1e-7. We use stochastic gradient descent (SGD) to update parameters and data augmentations to distort, jitter, crop, and resize inputs with size 160$\times$320.
The accuracy of SkyNet is shown in Table \ref{tab:training_accuracy}, where SkyNet C reaches the highest IoU (0.741) on the validation set. Therefore, we use SkyNet C as the proposed SkyNet for the following experiments.

\begin{table}[t]
\vspace{-0pt}
\caption{SkyNet validation  accuracy using the officially provided validation set with 1000 images}
\label{tab:training_accuracy}
\begin{center}
\begin{small}
\newcommand{\tabincell}[2]{\begin{tabular}{@{}#1@{}}#2\end{tabular}}
\begin{tabular}{|c|c|}
\hline
\textbf{DNN Model} & \textbf{IoU} \\
\hline
\tabincell{c}{SkyNet A} & \tabincell{c}{0.686} \\ 
\hline
\tabincell{c}{SkyNet B} & \tabincell{c}{0.703} \\ 
\hline
\tabincell{c}{SkyNet C} & \tabincell{c}{\textbf{0.741}} \\ 
\hline
\end{tabular}
\end{small}
\end{center}
\vskip -0.2in
\end{table}

\section{Experiments}

We demonstrate the capability of the proposed SkyNet for the low power object detection challenge in DAC-SDC 2019.
The evaluation is based on detection accuracy (IoU), inference throughput (FPS), and energy consumption (J). The calculation of final score is defined in~\cite{xu2018dac} as follows.

Assuming there are $I$ registered teams and $K$ images in the test set, the IoU score for team $i$, denoted as $R_{IoU_i}$, is calculated as: 
\begin{equation}
\label{eq:iou}
    R_{IoU_i} = \frac{ \sum\limits_{k=1}^{K} IoU_{i,k}}{K}
\end{equation}

For energy, $\bar{E}_{I}$ is denoted as the average energy consumption of all $I$ entries when performing DNN inference on the test dataset (Equation \ref{eq:E_avg}). The energy score of team $i$ ($ES_{i}$) is then computed using Equation \ref{eq:ES} relating to the ratio between average energy and the energy consumed by this team. $x$ in this equation is set to 2 and 10 for FPGA category and GPU category, respectively. 

\begin{equation}
\label{eq:E_avg}
    \bar{E}_{I} = \frac{ \sum\limits_{i=1}^{I} E_{i}}{I}
\end{equation}

\begin{equation}
\label{eq:ES}
    ES_{i} = max\{0, 1+0.2\times log_x \frac{\bar{E}_I}{E_i}  \}
\end{equation}

Eventually, the total score, denoted as $TS_{i}$, is calculated in Equation \ref{eq:TS} including both inference accuracy ($R_{IoU_i}$) and energy consumption ($ES_i$). 

\begin{equation}
\label{eq:TS}
    TS_{i} = R_{IoU_i} \times (1+ ES_i )
\end{equation}

\subsection{Result comparison}
The proposed SkyNet is deployed onto the given GPU and FPGA platforms. For GPU implementation, we keep all network parameters using 32-bit float data format; for the FPGA design, we quantize the feature maps and parameters to 9 bits and 11 bits, respectively, for better hardware performance.
The final results of the top-3 teams are listed in Table \ref{tab:rst_comp_gpu} and \ref{tab:rst_comp_fpga} \cite{2019DACSDC}. In total, 52 GPU teams and 58 FPGA teams participated worldwide creating a very intense competition. Our SkyNet design has successfully delivered the best inference accuracy and total score for both GPU and FPGA tracks. 

\begin{table*}[t]
\vspace{-8pt}
\caption{GPU final results from DAC-SDC'19 and '18 using the hidden test set with 50K images, evaluated by a TX2 GPU.}
\label{tab:rst_comp_gpu}
\begin{center}
\begin{small}
\newcommand{\tabincell}[2]{\begin{tabular}{@{}#1@{}}#2\end{tabular}}
\begin{tabular}{|c|c|c|c|c|c|}
\hline
\textbf{Team Name} & \textbf{Backbone} & \textbf{IoU} & \textbf{FPS} & \textbf{Power(W)} & \textbf{Total Score} \\
\hline
\hline
\multicolumn{6}{|c|}{Results from 2019} \\
\hline
\tabincell{c}{iSmart3-SkyNet (ours) \\ \cite{iSmart3-SkyNet}} & \tabincell{c}{\textbf{SkyNet}} & \textbf{0.731} & \textbf{67.33} & 13.50 & \textbf{1.504} \\ 
\hline
\tabincell{c}{Thinker \\ \cite{thinker}} & \tabincell{c}{ShuffleNet} & 0.713 & 28.79 & \textbf{8.55} & 1.442 \\ 
\hline
\tabincell{c}{DeepZS \\ \cite{DeepZS}} & \tabincell{c}{Tiny YOLO + \\ResNet-18
} & 0.723 & 26.37 & 15.12& 1.422 \\ 
\hline
\hline
\multicolumn{6}{|c|}{Results from 2018} \\
\hline
\tabincell{c}{ICT-CAS \\ \cite{ICT-CAS}} & \tabincell{c}{Tiny YOLO} & 0.698 & 24.55 & 12.58& 1.373 \\ 
\hline
\tabincell{c}{DeepZ \\ \cite{DeepZ}} & \tabincell{c}{Tiny YOLO} & 0.691 & 25.30 & 13.27 & 1.359 \\ 
\hline
\tabincell{c}{SDU-legend \\ \cite{SDU-Legend}} & \tabincell{c}{YOLOv2} & 0.685 & 23.64 & 10.31& 1.358 \\ 
\hline
\end{tabular}
\end{small}
\end{center}
\vskip -0.15in
\end{table*}

\begin{table*}[t!]
\vspace{-0pt}
\caption{FPGA final results in DAC-SDC'19 and '18 using the hidden test set with 50K images. Designs in 2019 are evaluated on a Ultra96 FPGA while designs in 2018 use a Pynq-Z1 FPGA.}
\label{tab:rst_comp_fpga}
\begin{center}
\begin{small}
\newcommand{\tabincell}[2]{\begin{tabular}{@{}#1@{}}#2\end{tabular}}
\begin{tabular}{|c|c|c|c|c|c|}
\hline
\textbf{Team Name} & \textbf{Backbone} & \textbf{IoU} & \textbf{FPS} & \textbf{Power (W)} & \textbf{Total Score} \\
\hline
\hline
\multicolumn{6}{|c|}{Results in 2019} \\
\hline
\tabincell{c}{iSmart3 (ours) \\ \cite{iSmart3}} & \tabincell{c}{\textbf{SkyNet}} & \textbf{0.716} & 25.05 & 7.26 & \textbf{1.526} \\ 
\hline
\tabincell{c}{XJTU\_Tripler \\ \cite{XJTU}} & \tabincell{c}{ShuffleNetV2} & 0.615 & 50.91 & 9.25 & 1.394 \\ 
\hline
\tabincell{c}{SystemsETHZ \\ \cite{systemsETHZ19}} & \tabincell{c}{SqueezeNet} & 0.553 & \textbf{55.13} & \textbf{6.69} & 1.318 \\ 
\hline
\hline
\multicolumn{6}{|c|}{Results in 2018} \\
\hline
\tabincell{c}{TGIIF \\ \cite{TGIIF}} & \tabincell{c}{SSD} & 0.624 & 11.96 & 4.20& 1.267 \\ 
\hline
\tabincell{c}{SystemsETHZ \\ \cite{systemsETHZ}} & \tabincell{c}{SqueezeNet} & 0.492 & 25.97 & 2.45 & 1.179 \\ 
\hline
\tabincell{c}{iSmart2 \\ \cite{iSmart2}} & \tabincell{c}{MobileNet} & 0.573 & 7.35 &2.59& 1.164 \\ 
\hline
\end{tabular}
\end{small}
\end{center}
\vskip -0.2in
\end{table*}

\section{Conclusions and Discussions}
In this paper, we proposed SkyNet, a lightweight DNN developed following a bottom-up design approach specializing in low power object detection. The proposed design was demonstrated on the 56th IEEE/ACM Design Automation Conference System Design Contest (DAC-SDC) and won the first place award for both GPU and FPGA tracks.
The great success in DAC-SDC also indicated that the proposed bottom-up DNN design approach is effective for enhancing the performance of object detection on embedded GPUs and FPGAs. The method can be extended to other edge devices with appropriate latency and resource modeling. Bundles can be enumerated and evaluated on targeted devices, and with the specific performance targets, DNN can be grown by stacking Bundles following the guidance of search algorithms. In our case, we used SCD to approach performance targets, and other search algorithms are also feasible. The proposed design method can also be extended to other DNN-related edge applications, such as classification, recognition, object tracking, etc.

\section{ Acknowledgments}
This work was partly supported by the IBM-Illinois Center for Cognitive Computing System Research (C$^3$SR) -- a research collaboration as part of IBM AI Horizons Network. 
The authors would like to express their deep gratitude for additional team members in team \textit{\textbf{iSmart3-SkyNet}} (GPU track): Sitao Huang, Bowen Cheng, Yunchao Wei, and team members in team \textit{\textbf{iSmart3}} (FPGA track): Yao Chen, Xingheng Liu, Sitao Huang.

\bibliographystyle{aaai}
\small
\bibliography{ref}
\end{document}